\documentclass[runningheads]{llncs}

\usepackage{eccv}
\usepackage{multirow}
\usepackage{booktabs}
\usepackage{tikz}
\usepackage{bm}
\usepackage{xurl}
\usepackage{adjustbox}
\usepackage{wrapfig}
\usepackage{microtype}
\usepackage{array}
\usepackage{tabularx}

\definecolor{cvprblue}{rgb}{0.21,0.49,0.74}
\usepackage[breaklinks,colorlinks,allcolors=cvprblue]{hyperref}

\title{Neural Reconstruction of LiDAR Point Clouds under Jamming Attacks via Full-Waveform Representation and Simultaneous Laser Sensing}
\titlerunning{Neural Reconstruction of LiDAR point cloud under Jamming}
\authorrunning{R. Yoshida et al.}

\author{
    Ryo Yoshida\inst{1} \and 
    Takami Sato\inst{1} \and 
    Wenlun Zhang\inst{1} \and 
    Yuki Hayakawa\inst{1} \and 
    Shota Nagai\inst{1} \and 
    Takahiro Kado\inst{2} \and 
    Taro Beppu\inst{2} \and 
    Ibuki Fujioka\inst{2} \and 
    \\Yunshan Zhong\inst{3} \and 
    Kentaro Yoshioka\inst{1}
}

\institute{
    Keio University, Yokohama, Japan \\
    \email{\{ryo-yoshida, wenlun\_zhang, hykwyuk, n-shota, kyoshioka47\}@keio.jp, takami.sato@yos.elec.keio.ac.jp} \and
    Sony Semiconductor Solutions, Atsugi, Japan \\
    \email{\{Takahiro.Kado, Taro.Beppu, Ibuki.Fujioka\}@sony.com} \and
    Hainan University, Haikou, China \\
    \email{viperzhong@163.com}
}

\begin{document}
\maketitle

\begin{abstract}
LiDAR sensors are critical for autonomous driving perception, yet remain vulnerable to spoofing attacks. Jamming attacks inject high-frequency laser pulses that completely blind LiDAR sensors by overwhelming authentic returns with malicious signals. We discover that while point clouds become randomized, the underlying full-waveform data retains distinguishable signatures between attack and legitimate signals. In this work, we propose PULSAR-Net, capable of reconstructing authentic point clouds under jamming attacks by leveraging previously underutilized intermediate full-waveform representations and simultaneous laser sensing in modern LiDAR systems. PULSAR-Net adopts a novel U-Net architecture with axial spatial attention mechanisms specifically designed to identify attack-induced signals from authentic object returns in the full-waveform representation. To address the lack of full-waveform representations in existing LiDAR datasets under jamming attacks, we introduce a physics-aware dataset generation pipeline that synthesizes realistic full-waveform representations under jamming attacks. Despite being trained exclusively on synthetic data, PULSAR-Net achieves reconstruction rates of 92\% and 73\% for vehicles obscured by jamming attacks in real-world static and driving scenarios, respectively.

\keywords{Autonomous Driving \and LiDAR \and Spoofing Attack}
\end{abstract}

\section{Introduction}

\begin{figure}[ht!]
\centering
\label{fig:1}
\includegraphics[width=\linewidth]{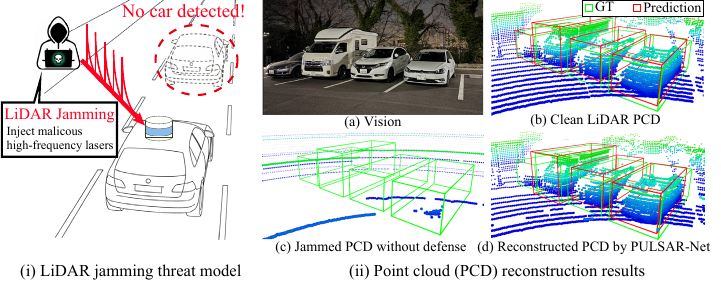}
\vspace{-0.22in}
\caption{
Reconstruction demo of our PULSAR-Net from LiDAR sensing under a jamming attack. Our PULSAR-Net can effectively recover the point cloud (d), which is faithful to the original one (b), by leveraging the intermediate full-waveform representation and simultaneous sensing feature available in modern LiDARs. }
\label{fig:poc}
\vspace{-0.15in}
\end{figure}

LiDAR (Light Detection and Ranging) sensors have become indispensable components of autonomous driving systems, providing precise 3D environmental perception critical for safe navigation. By emitting laser pulses and measuring their time-of-flight (ToF), LiDAR generates detailed point clouds that enable vehicles to detect obstacles, localize themselves, and make informed driving decisions. However, this reliance on active sensing introduces a fundamental security vulnerability: LiDAR systems are vulnerable to spoofing attacks that inject malicious infrared laser signals at the same wavelength as the sensor's legitimate lasers~\cite{shin2017illusion, cao2019adversarial}.
Recent research has demonstrated various spoofing attacks that can deceive autonomous vehicles by creating phantom objects or concealing real obstacles~\cite{cao2023you, sato2024lidar}. Among these threats, jamming attacks represent one of the most severe forms, as no prior work~\cite{sato2024lidar, sato2025on, suzuki2025lab} has developed effective countermeasures against them. In a jamming attack, the adversary floods the LiDAR sensor with rapid laser pulses, completely randomizing the received point cloud~\cite{sato2024lidar, sato2025on}. This disruption renders the victim vehicle unable to perceive its surrounding environment, potentially leading to fatal consequences, e.g., colliding with a leading vehicle as depicted in Fig.~\ref{fig:poc} (i). While modern LiDAR manufacturers have implemented several countermeasures, none of the existing defense mechanisms have effectively mitigated jamming attacks~\cite{sato2024lidar}. These risks highlight the urgent need for effective countermeasures against LiDAR jamming in autonomous driving systems.

In this work, we challenge this assumption by advocating that modern LiDAR systems already capture rich but abandoned information that enables discrimination between legitimate and malicious lasers. We identify two meaningful information sources: full-waveform representation and simultaneous laser sensing. Full-waveform representation, which records the complete temporal profile of returned laser intensity over the entire sensing period, contains subtle but remarkable signatures that distinguish authentic object reflections from attack-induced signals. Simultaneous laser sensing is a common feature in recent LiDARs to measure multiple points simultaneously, which enabling us to identify attack signals. Because the attacker's laser beam diffuses over an area larger than the LiDAR's diameter, the resulting attack signal produces identical full-waveforms across simultaneously sensing receivers. In contrast, legitimate returns vary based on the specific objects reflecting them. To leverage the findings, we design PULSAR-Net (Point cloud Un-jamming via LiDAR Signal Adaptive Reconstruction), the first successful defense against jamming attacks on LiDAR systems. We adopt a novel U-Net-based architecture with axial spatial attention, explicitly designed for attack signal segmentation in full-waveform representation, to effectively capture the structure of attack pulses in simultaneously sensed full-waveform data.

We first design a realistic synthesis methodology to generate full waveforms under jamming attacks from existing point cloud datasets. Our goal is to enable PULSAR-Net to generalize effectively to real-world LiDAR point clouds despite being trained solely on synthesized data derived from benign point clouds. This approach is necessary because collecting LiDAR data under actual jamming attacks is generally impractical, as deploying such attacks across diverse real-world scenarios is highly challenging. For this spurpose we reconstruct the intermediate full-waveform representation from post-processed point clouds given known LiDAR specifications. Building on these synthesized waveforms and the scan pattern of simultaneous laser sensing, we apply a jamming attack synthesis methodology that injects high-frequency attack pulses with timing randomization to reproduce realistic jamming conditions.

To validate the effectiveness of PULSAR-Net, we conduct a large-scale evaluation on the synthesized datasets generated from popular LiDAR datasets, KITTI and nuScenes. We further validate the real-world generalizability of PULSAR-Net by integrating it with a production-ready LiDAR we developed with a major LiDAR supplier to stream full-waveform data. Despite being trained exclusively on synthetic data, PULSAR-Net demonstrates strong generalization to physical attack scenarios, as illustrated in Fig.~\ref{fig:poc} (ii). These results establish PULSAR-Net as the first practical defense mechanism capable of recovering point clouds under LiDAR jamming attacks. This work makes the following contributions:

\begin{itemize}
    \item [$\bullet$] We design PULSAR-Net, the first successful defense against jamming attacks on LiDAR systems, featuring a U-Net-based architecture with axial spatial attention specifically designed for attack-signal segmentation in the full-waveform representation. 
    
    \item [$\bullet$] We develop a physics-based synthesis pipeline that generates realistic full-waveform data under jamming attacks from existing point cloud datasets. 
    
    \item [$\bullet$] We demonstrate that PULSAR-Net achieves a recovery rate of $\geq$82\% for vehicles obscured by jamming attacks on our generated dataset, representing an $\geq$2-fold improvement over a naive baseline approach on our datasets.
    
    \item [$\bullet$] In collaboration with a major supplier, we integrated PULSAR-Net into a production-ready LiDAR with full-waveform access. Despite training solely on simulated data, it achieves reconstruction rates of 92\% and 73\% in real-world static and driving scenarios, respectively.
    
\end{itemize}

\noindent Our code and datasets will be released when published.

\section{Related Work}
\vspace{-0.1in}
\subsection{LiDAR Spoofing Attacks}
\vspace{-0.05in}

LiDAR spoofing attacks exploit the fundamental principle of active laser sensing by injecting malicious laser pulses that interfere with legitimate measurements~\cite{petit2015remote, shin2017illusion, cao2019adversarial, jiachen2020towards, hallyburton2022security, jin2022pla, sato2024lidar, cao2023you, cao2021invisible}. An attacker equipped with an infrared laser source operating at the same wavelength as the target LiDAR can inject false signals that are indistinguishable from legitimate ones. HFR attack~\cite{sato2024lidar,sato2025on} is known as one of the most effective LiDAR spoofing attacks, which emits laser pulses at a frequency significantly higher than the LiDAR's own pulse rate, saturating the sensor's receiver. This results in complete randomization of the perceived point cloud, as nearly every laser pulse fired by the LiDAR receives multiple false returns from the attacker's continuous emission, overwhelming authentic object reflections as illustrated in Fig.~\ref{fig:full_wave} (i).
Modern LiDARs have several functionalities that can serve as mitigations, originally implemented to mitigate interference from other LiDARs, e.g., pulse timing randomization~\cite{cao2019adversarial, cao2023you} and pulse fingerprinting~\cite{shin2017illusion}. 
However, prior research~\cite{sato2025on} demonstrates that none of these mitigations are sufficient to prevent serious consequences such as failing to detect pedestrians on the road. 
The fundamental challenge lies in the nature of the jamming attack: once the point cloud is compromised, the resulting data becomes nearly random noise with no informative structure that could enable recovery of the original scene geometry. The attack-induced points are statistically indistinguishable from legitimate returns when examined solely at the point cloud level. This has led to a belief that defending against jamming attacks through post-processing is virtually impossible.

\vspace{-0.1in}
\subsection{Full-Waveform Representation in LiDAR}\label{sec:full_waveform}
\vspace{-0.05in}

\begin{figure}[t!]
    \centering
    \includegraphics[width=\linewidth]{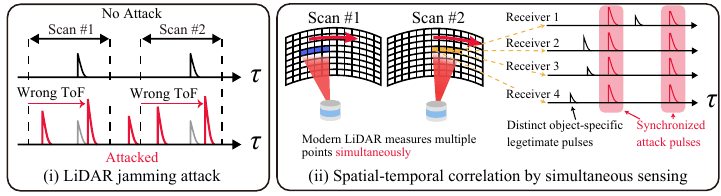}
    \caption{Conceptual illustration of a LiDAR jamming attack and simultaneous sensing.}
    \label{fig:full_wave}
    \vspace{-0.25in}
\end{figure}

While jamming attacks, such as HFR attack, can make the point cloud completely random, this does not mean there is no way to obtain information from the legitimate laser reflection, because the legitimate laser reflection does actually reach the LiDAR sensor, albeit typically weaker than the attack pulses. For each point in the point cloud, the LiDAR obtains a time-resolved full waveform from the returning laser light, and then a digital signal processor (DSP) pipeline detects a peak in the waveform~\cite{scheuble2025lidar, goudreault2023lidar}. The peak is generally one of the strongest attack pulses as shown in Fig.~\ref{fig:full_wave} (i). Critically, only the final point cloud is typically stored and used for downstream perception tasks. The intermediate full-waveform data, which contains rich temporal information, is discarded after peak detection to reduce data storage, bandwidth, and latency requirements. A recent study~\cite{scheuble2025lidar} finds that the full-waveform representation can fit a tokenization of Transformer~\cite{vaswani2017attention} and demonstrates the performance improvements, particularly in foggy conditions.

\vspace{-0.1in}
\subsection{Simultaneous Laser Sensing}
\vspace{-0.05in}
\label{sec:scan_pattern}

Modern LiDAR systems employ a critical design feature: they fire lasers from multiple emitters simultaneously and receive them within the same time \cite{sato2024lidar}.
Fig.~\ref{fig:full_wave} (ii) illustrates how LiDAR scans several adjacent points at the same time with multiple laser emitters and receivers. This creates strong spatial-temporal correlations in the full-waveform data, allowing us to distinguish the attack laser pulses from legitimate laser reflections. Specifically, the same pattern of the attack pulses will be observed in multiple full waveforms received at the same time as shown in Fig.~\ref{fig:full_wave} (ii). The attacker's laser is emitted from outside the LiDAR and is dispersed beyond the diameter of the LiDAR, making it impossible to selectively attack a specific receiver. On the other hand, the legitimate laser is emitted inside the LiDAR, and thus each receiver can receive the corresponding laser due to the directness of light; i.e., the detected laser patterns differ based on the object hit by each laser beam. 
Based on this observation, one might consider a naive approach: remove attack pulses by canceling out patterns that commonly appear across simultaneously received full-waveforms. However, as we discuss in ~\S\ref{sec:main_result}, this simple method fails in practice—it cannot entirely remove attack pulses and produces a high number of false positives.

To fully utilize the information in the full-waveform representation and the simultaneous laser sensing, we design PULSAR-Net, a novel U-Net architecture specifically designed for attack signal segmentation in the full-waveform representation to capture the highly correlated attack pulses within the same scan. PULSAR-Net aims to capture other unique characteristics that legitimate laser reflections do not exhibit. For example, attack lasers typically exhibit higher and more consistent peak intensities as they are direct laser emissions rather than diffuse reflections from objects, and temporal patterns that appear simultaneously across multiple receivers, while legitimate reflections arrive at different times depending on object distances. By capturing these characteristics, we demonstrate that it is possible to recover the original point cloud with a high success rate, a task previously considered impossible.

\vspace{-0.08in}
\section{Full-Waveform Jamming Attack Simulation} \label{sec:data_gen}
\vspace{-0.03in}
To effectively train our PULSAR-Net, we first design a methodology to generate attacked full-waveform datasets with simultaneous laser-sensing patterns from existing LiDAR point cloud data. Ideally, the data should be collected in the real world without any synthesis, but gathering full-waveform data and annotating it at scale is highly costly. However, data collection requires significant effort to cover a wide variety of LiDARs, scenes, attack setups, and victim objects. Moreover, setting up jamming attacks at every data collection point is practically infeasible. To address this, we simulate the intermediate full-waveform representation from given LiDAR specifications and point cloud data in existing LiDAR datasets such as KITTI~\cite{geiger2013vision}, nuScenes~\cite{caesar2020nuscenes}, and Waymo Open Dataset~\cite{waymo_dataset}, which include only the final point cloud outputs $(x, y, z, \text{intensity})$ without the full-waveform representations essential for our defense mechanism. Our goal is to achieve sufficient real-world generalizability while training only on simulated data.

\noindent\textbf{Notation.}
We denote the full-waveform tensor as $\bm{\mathcal{A}} \in \mathbb{R}^{H \times W \times D}$, where $H$, $W$, and $D$ denote the number of channels (vertical angles), horizontal resolution (azimuth samples), and temporal bins (full-waveform bin length), respectively. $\bm{\mathcal{A}}$ represents a single LiDAR scan of 360$^{\circ}$. For each measurement direction $(i, j)$ where $i \in \{1, \ldots, H\}$ and $j \in \{1, \ldots, W\}$, the LiDAR captures a full waveform $\bm{\mathcal{A}}_{ij} \in \mathbb{R}^D$ over time bins $t \in \{0, \ldots, D-1\}$. We denote the discrete scan timestamp of a single 360$^\circ$ scan as $\Theta = \{0, 1, \ldots, \tau_{\text{max}}-1\}$.
\vspace{-0.1in}
\subsection{Full-Waveform Representation Modeling}
\vspace{-0.05in}
Given benign point cloud data from an existing LiDAR dataset, we model its full-waveform representation. The point cloud data only provides the location of the strongest return at each measurement $(i, j)$. A naive approach would be to assign a non-zero intensity value in $\bm{\mathcal{A}}_{ij}$ at the corresponding time bin, and set the others to zero. However, this digital approach fails to capture the analog characteristics of the reflected laser pulse. Following prior work~\cite{yoshioka201820}, we adopt a Gaussian pulse model $p(t) = \exp\left(-\frac{t^2}{2\sigma^2}\right)$ to simulate realistic reflected pulse returns. We model the full-waveform tensor $\bm{\mathcal{L}} \in \mathbb{R}^{H \times W \times T}$ of a given point cloud as
  $\mathcal{L}_{ijt} = \gamma \, i_{ij}\cdot p(t - 2d_{ij}/c_0),$
where $\gamma$ is the pulse amplitude scaling factor to convert the unit Gaussian pulse to an appropriate amplitude for the target LiDAR (we used $\gamma$=$12$ for the KITTI dataset and $\gamma$=$0.156$ for the nuScenes dataset to surpass the legitimate pulses sufficiently), 
$d_{ij}$ and $i_{ij}$ are the distance and intensity at measurement direction $(i,j)$, and $c_0$ is the light speed.
\vspace{-0.1in}
\subsection{Simultaneous Laser Sensing Modeling}
\vspace{-0.05in}
As described in~\S\ref{sec:scan_pattern}, we leverage the unique characteristic of modern LiDARs that emit and sense multiple lasers simultaneously. To model this behavior, we introduce a scan timing matrix $\bm{\mathcal{T}} \in \Theta^{H \times W}$ whose element $\mathcal{T}_{ij} \in \Theta$ contains the scan start timestamp of each measurement direction $(i, j)$. Two measurements $(i, j)$ and $(s, k)$ are scanned simultaneously if and only if $\mathcal{T}_{ij} = \mathcal{T}_{sk}$.
\vspace{-0.1in}
\subsection{Jamming Attack Modeling for Full Waveforms}
\vspace{-0.05in}

We model the attacked full-waveform tensor as the superposition of the benign signal, the attack signal, and the noise. Let $\bm{\mathcal{L}} \in \mathbb{R}^{H \times W \times D}$ denote the benign full-waveform tensor, $\bm{\mathcal{F}}: \Theta^{H \times W} \rightarrow \mathbb{R}^{H \times W \times D}$ denote the attack pulse generation function parameterized by the scan timing matrix, and $\bm{\mathcal{N}} \in \mathbb{R}^{H \times W \times D}$ denote the observation noise. The attacked full-waveform tensor $\check{\bm{\mathcal{A}}} \in \mathbb{R}^{H \times W \times D}$ is  given by
  $\check{\bm{\mathcal{A}}} = \bm{\mathcal{L}} + \bm{\mathcal{F}}(\bm{\mathcal{T}}) + \bm{\mathcal{N}}$.
The function $\bm{\mathcal{F}}(\bm{\mathcal{T}})$ ensures that measurements with identical scan timestamps receive the same attack waveform, reflecting the simultaneous laser sensing. For each measurement direction $(i, j)$, the attack pulse $\bm{\mathcal{F}}(\bm{\mathcal{T}})_{ijt}$ is constructed as a periodic train of Gaussian pulses:
$\bm{\mathcal{F}}(\bm{\mathcal{T}})_{ijt} = \gamma_A \sum_{k=1}^{K} p(t - (\tau_k + \epsilon_k))$, 
where $K = \lfloor T_{\max} / \delta_{\rm attack} \rfloor$ is the number of attack pulses within the observation window, $T_{\max}$ is the total time duration of the full waveform, $\delta_{\rm attack}$ is the attack pulse interval (e.g., $\delta_{\rm attack} = 1$ $\mu$s for a 1 MHz attack pulse frequency), $\gamma_A$ is the attack pulse amplitude, and $\tau_k = k \cdot \delta_{\rm attack}$ is the nominal timing of the $k$-th attack pulse. 
We add timing jitter $\epsilon_k \sim \mathcal{U}(-20 \text{ ns}, 20 \text{ ns})$ to each pulse timing $\tau_k$.
This design can model more general jamming attacks than current state-of-the-art attacks, such as HFR~\cite{sato2024lidar} and A-HFR~\cite{sato2025on}attacks, which inject periodic attack pulses that may be predictable to the defense side. The noise term $\epsilon_k$ disrupts the structure of the attack pulses, enabling us to simulate a realistic, randomized jamming attack.

\vspace{-0.1in}

\section{PULSAR-Net: Neural LiDAR PCD Reconstruction}
\vspace{-0.05in}
\begin{figure}[t!]
    \centering
        \includegraphics[width=\linewidth]{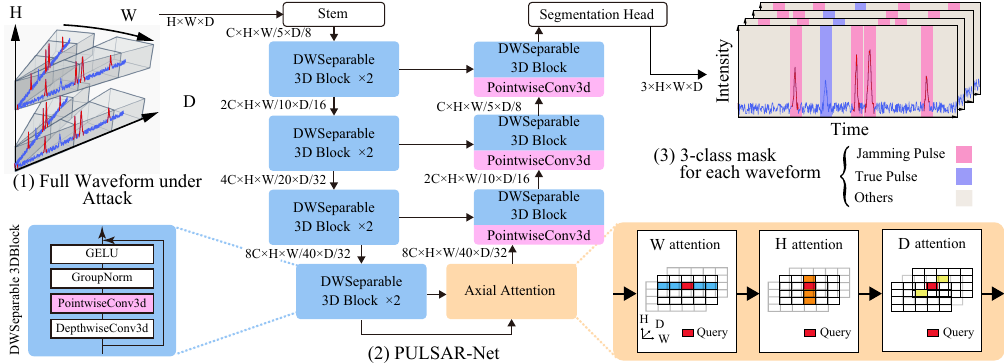}
        \vspace{-0.2in}
        \caption{Overview of PULSAR-Net, a U-Net-inspired architecture using depthwise-separable 3D convolutions for efficient, lightweight feature extraction. It incorporates axial attention to model both spatial and temporal dependencies within full-waveforms. }
        \label{fig:arch}
        \vspace{-0.18in}
\end{figure}
To fully leverage the full-waveform representation (\S\ref{sec:full_waveform}) and the simultaneous laser sensing (\S\ref{sec:scan_pattern}), we design PULSAR-Net, a novel U-Net-based neural network architecture with axial spatial attention specifically designed for attack signal segmentation in full-waveform data as depicted in Fig.~\ref{fig:arch}. Our depthwise-separable 3D convolutions and axial attention, can capture LiDAR-agnostic physical principles rather than dataset-specific artifacts, enabling robust defense performance across diverse LiDAR systems without requiring hardware-specific retraining.

\vspace{-0.1in}
\subsection{Segmentation-based Full-Waveform Denoising} \label{sec:seg_head}
\vspace{-0.05in}
We formulate the LiDAR point cloud reconstruction as a segmentation-based denoising in full-waveform representation. For each measurement, we predict three classes: (1) background (no reflection), (2) legitimate reflection from an object, and (3) attack pulse. Our goal is to train a denoising model $f_{D}(\cdot)$ that can accurately predict the binary segmentation mask $\bm{\mathcal{M}}_{attack}$ corresponding to the malicious attack pulses in the full waveforms. We apply the mask to suppress the attack pulses in the full waveform as follows:
$\hat{\bm{\mathcal{A}}} = \check{\bm{\mathcal{A}}} \odot (1 - \bm{\mathcal{M}}_{attack})$.
We obtain the reconstructed point cloud by taking the peak of each denoised full waveform. This design reduces computational cost and even improves performance. A naive approach is to predict the denoised full waveform via regression. However, this approach underperformed since information was lost during resizing before feeding to the reconstruction model. In contrast, our masking-based approach is more robust because resizing is applied to the mask, preserving rich information of the full waveform in the high-resolution original space.

\vspace{-0.1in}
\subsection{Model Architecture}
\vspace{-0.05in}
We design a dedicated 3D U-Net architecture that operates directly on LiDAR waveform data, as illustrated in Fig.~\ref{fig:arch}, to satisfy the two key requirements: \textbf{i)} the model should fully exploit both spatial and temporal information to accurately identify malicious peaks within the waveform data accurately \textbf{ii)} the model should be computationally efficient to enable real-time signal processing, thereby improving its practical applicability. We adopt a 3D U-Net architecture that operates directly on LiDAR waveform data, as illustrated in Fig.~\ref{fig:arch}. The encoder is composed of multiple depthwise-separable 3D convolutional blocks and progressive downsampling layers to extract multi-scale spatial--temporal features. The decoder mirrors the encoder with corresponding upsampling stages, where skip connections integrate high-resolution details with semantically rich features from the encoder. Finally, a segmentation head predicts the attack mask for each voxel, enabling precise separation between malicious and clean waveform peaks. To further strengthen these two design objectives, we incorporate two specialized blocks into the network.

\noindent\textbf{Depthwise-Separable 3D Convolution.} 
Since full-waveform data can be large, directly applying standard 3D convolutions becomes computationally prohibitive.
A typical LiDAR waveform with dimensions (32, 1800, 800)—representing vertical channels, horizontal resolution, and waveform bins—exceeds the memory limits of even high-end GPUs.
To address this and enhance computational efficiency while retaining the representational capacity of 3D convolutions, we adopt depthwise-separable 3D convolutions inspired by DW-3D~\cite{ye20193d}. Unlike standard 3D convolutions that operate simultaneously across spatial and channel dimensions, depthwise-separable 3D convolutions factorize the operation into two stages: \textbf{1)} a depthwise 3D convolution applied independently to each input channel, and \textbf{2)} a pointwise $1 \times 1 \times 1$ convolution that integrates inter-channel information.

Specifically, a standard 3D convolution layer has 
$(k_{x} \cdot k_{y} \cdot k_{z} \cdot C_{in} \cdot C_{out})$ parameters,
where $k_{x}$, $k_{y}$, and $k_{z}$ denote the kernel sizes, and $C_{in}$ and $C_{out}$ represent the input and output channel dimensions, respectively. In contrast, the parameter count of a depthwise-separable 3D convolution is reduced to $(k_{x} \cdot k_{y} \cdot k_{z} \cdot C_{in} + C_{in} \cdot C_{out})$.
\noindent For instance, given a denoising model with kernel size $(3, 5, 7)$ and $C_{out} = 64$, the total parameters and operations can be reduced by over 97\%. This design preserves the representational capacity of 3D convolutions while drastically lowering both the parameter count and computational cost, making the model well-suited for real-time LiDAR deployment.

\noindent\textbf{Axial Spatial-Temporal Attention.} 
While depthwise-separable 3D convolutions in the U-Net structure capture local spatial–temporal dependencies, they remain limited in modeling global waveform correlations. To overcome this limitation, we introduce an axial attention module that applies self-attention sequentially along the azimuth, elevation, and temporal (waveform) axes. Compared with standard 3D attention, which jointly processes all tokens and incurs cubic complexity, axial decomposition captures long-range dependencies within each dimension at a fraction of the computational cost. 
This axial design is particularly well-suited for exploiting simultaneous laser sensing patterns. When the LiDAR fires multiple lasers simultaneously along the azimuth or elevation axes, legitimate returns exhibit different waveform patterns depending on the objects each laser beam encounters. In contrast, attack signals injected from outside the LiDAR manifest as identical waveform patterns across all simultaneously-sensed channels, as the attacker cannot selectively target individual receivers, as discussed in ~\S\ref{sec:scan_pattern}. By applying attention along the azimuth and elevation axes independently, our architecture can efficiently compare waveforms across simultaneously-sensed channels and detect this characteristic signature of attack signals' identical patterns appearing across spatial dimensions where legitimate signals should differ.

\noindent\textbf{Loss Function Design.}
To handle the unbalanced of the classes, we use a weighted cross entropy loss, $\mathcal{L}_{\text{WCE}} = - \sum_{c \in \{0,1,2\}} w_c \sum_{i} y_{i,c} \log p_{i,c}$, where $(w_0,w_1,w_2)$ = $(0.1, 0.3, 0.6)$. With this loss and Dice loss~\cite{sorensen1948method}, we train the PULSAR-Net with the loss function $\mathcal{L}_{\text{WCE}} + 0.5 \cdot \lambda_{\text{Dice}}$.
\vspace{-0.1in}
\section{Evaluation}
\vspace{-0.05in}

We evaluate PULSAR-Net on synthesized jamming-attack datasets generated using the methodology in~\S\ref{sec:data_gen} and on a real-world attack case study. Our evaluation targets two questions: (i) can the model recover the original point geometry at the point level, and (ii) does the recovered point cloud preserve downstream perception performance for 3D object detection.

\vspace{-0.1in}
\subsection{Experimental Setup}
\vspace{-0.05in}
\noindent\textbf{Evaluation Dataset.} We construct our evaluation datasets from KITTI~\cite{geiger2013vision} and  nuScenes~\cite{caesar2020nuscenes} datasets. We synthesize the jamming attack-impacted data according to the methodology described in~\S\ref{sec:data_gen}. For each dataset, we extracted 1,100 scenes: 1,000 for training and 100 for testing. We inject the jamming attack effects into $\pm$45$^{\circ}$ in front of the victim with the target LiDAR. We call the datasets as KITTI-J and nuScenes-J, respectively.

\noindent\textbf{LiDAR Specifications.} We always use (32, 1800, 800), i.e., 32 vertical channels, 0.2$^{\circ}$ horizontal resolution, and 800 full-waveform dimensions, as the input from LiDAR. This specification is consistent with HDL-32E~\cite{HDL32E} LiDAR used in nuScenes datasets. The KITTI dataset uses HDL-64E~\cite{HDL64E}, which has more vertical channels and horizontal resolution. We downsample it to 32 vertical channels and 0.2$^{\circ}$ horizontal resolution. We note that this downsampling is only applied when training the attack mask as described in~\S\ref{sec:seg_head} and the final point-cloud dimension will not be changed. For the simultaneous laser sensing, we assume that the consecutive horizontal points are scanned at the same time, according to the modern design of LiDAR, such as Huawei L107~\cite{L107} and Hesai AT128P~\cite{AT128P}.

\noindent\textbf{Metrics.} We evaluate the denoising capability using two metrics: point-level recovery accuracy and object detector-level mAP. \textit{Point-level recovery accuracy} measures whether each reconstructed point in the 90$^\circ$ attack area is located within 0.5 m of the corresponding original benign point. Following LiDAR physics principles, this 0.5 m threshold is evaluated along the laser direction from the LiDAR origin, meaning the accuracy is calculated independently for each laser emission.
\textit{Object detector-level mAP} is a metric to evaluate the autonomous driving application-level defense capability. We calculate the mAP of all objects in the attack area, the 90$^\circ$ area in front of the victim.
We use PointPillars~\cite{lang2019pointpillars}, CenterPoint~\cite{yin2021center} and PV-RCNN~\cite{shi2020pv}, implemented in mmdetection3d~\cite{mmdetection3d}. We employed these models pre-trained on the KITTI and nuScenes datasets to evaluate KITTI-J and nuScenes-J, respectively.

\noindent\textbf{Baseline Methods.}
We compare our approach against two baselines: naive average subtraction and the Neural DSP~\cite{scheuble2025lidar}. Naive average subtraction simply computes the mean of simultaneously sensed full waveforms and subtracts this average from each individual waveform. We expect this averaged waveform to preserve the attack pulses while canceling out legitimate laser reflections, since the timing of legitimate peaks varies across lasers, as discussed in~\S\ref{sec:scan_pattern}. Neural DSP is a transformer-based architecture that directly processes full waveforms rather than point cloud data. Although it was not designed to handle attack-compromised data, we include it to demonstrate that our attack-specific reconstruction design outperforms the mere integration of full-waveform information.

\subsection{Results} \label{sec:main_result}

\begin{table}[t!]
  \footnotesize
  \centering
  \caption{
  \textit{Point-level recovery accuracy} of PULSAR-Net (Ours) and the baselines (Naive Average Subtraction and Neural DSP) on the nuScenes-J and KITTI-J datasets. $\phi$ is the number of simultaneous sensing lasers.
  }
  \vspace{-0.1in}
  \label{tab:accuracy}
  \setlength{\tabcolsep}{0.6pt}
  \setlength{\aboverulesep}{0pt}
  \setlength{\belowrulesep}{0pt}
  
  \begin{tabular}{c cc cc cc cc}
    \toprule
    & \multicolumn{2}{c}{\textbf{Ours}} & \multicolumn{2}{c}{Naive Average Sub.} & \multicolumn{2}{c}{Neural DSP~\cite{scheuble2025lidar}} & \multicolumn{2}{c}{No Defense} \\
    \cmidrule(lr){2-3} \cmidrule(lr){4-5} \cmidrule(lr){6-7} \cmidrule(lr){8-9}
    \shortstack[l]{$\phi$} & nuScenes-J & KITTI-J & nuScenes-J & KITTI-J & nuScenes-J & KITTI-J & nuScenes-J & KITTI-J \\
    \midrule
    1   & 67.95 & 73.21 & \multicolumn{1}{c}{-} & \multicolumn{1}{c}{-} & 4.57 & 3.92 & 3.37 & 2.92 \\
    4   & 82.66 & 89.63 & 38.22 & 28.6  & 6.21 & 21.4 & 3.36 & 2.87 \\
    5   & 86.13 & \textbf{89.90}  & \textbf{38.26} & 30.33 & 9.56 & 26.39 & 3.39 & 2.92 \\
    10  & 85.95 & 89.86 & 37.77 & 35.93 & 36.67 & \textbf{29.47} & 3.39 & \textbf{2.93} \\
    25  & 86.35 & 89.62 & 36.91 & 42.28 & 46.35 & 21.03 & \textbf{3.42} & 2.90  \\
    55  & 86.47 & 89.85 & 35.5  & \textbf{45.44} & \textbf{48.87} & 13.76 & 3.41 & 2.92 \\
    110 & 86.32 & 89.44 & 32.6  & 35.21 & 47.84 & 13.44 & 3.39 & 2.82 \\
    225 & \textbf{86.75} & 89.56 & 34.89 & 16.36 & 38.93 & 13.30 & 3.35 & 2.85 \\
    \bottomrule
  \end{tabular}
  \vspace{-0.24in}
\end{table}

\begin{table}[t]
  \centering
  \caption{
    \textit{Object detector-level mAP} of PULSAR-Net (Ours) and baselines on the nuScenes-J and KITTI-J datasets. $\phi$ denotes the number of simultaneous sensing lasers.
  }
  \label{tab:map}

  \footnotesize
  \setlength{\aboverulesep}{0pt}
  \setlength{\belowrulesep}{0pt}
  \begin{tabular*}{\columnwidth}{@{\extracolsep{\fill}}clcc|cc}
    \toprule
    & & \multicolumn{2}{c|}{\textbf{nuScenes-J (mAP)}} 
      & \multicolumn{2}{c}{\textbf{KITTI-J (mAP)}} \\
    $\phi$ & \textbf{Method} 
      & PointPillars 
      & CenterPoint 
      & PointPillars 
      & PV-RCNN \\
    \midrule
    
    \multirow{4}{*}{25} 
     & No Defense & 0.00 & 0.00 & 0.00 & 0.00 \\
     & Naive Average Sub.  & 1.94 & 0.00 & 21.81 & 41.06 \\
     & Neural DSP~\cite{scheuble2025lidar} & 20.09 & 4.07 & 4.32 & 6.25 \\
     & \textbf{Ours} & \textbf{64.03} & \textbf{59.08} & \textbf{48.33} & \textbf{56.06} \\
    \midrule

    \multirow{4}{*}{110} 
     & No Defense & 0.00 & 0.00 & 0.00 & 0.00 \\
     & Naive Average Sub.  & 2.56 & 4.04 & 16.36 & 32.94 \\
     & Neural DSP~\cite{scheuble2025lidar} & 37.70 & 26.51 & 0.57 & 1.43 \\
     & \textbf{Ours} & \textbf{65.52} & \textbf{59.00} & \textbf{48.10} & \textbf{55.74} \\
    \midrule

    \multirow{4}{*}{225} 
     & No Defense & 0.00 & 0.00 & 0.00 & 0.00 \\
     & Naive Average Sub. & 4.09 & 9.11 & 0.14 & 2.34 \\
     & Neural DSP~\cite{scheuble2025lidar} & 31.05 & 22.49 & 0.34 & 0.63 \\
     & \textbf{Ours} & \textbf{64.15} & \textbf{59.25} & \textbf{48.95} & \textbf{56.67} \\
    \midrule

     & No Attack & 69.87 & 72.72 & 57.51 & 74.83 \\
    \bottomrule
  \end{tabular*}
  \vspace{-0.18in}
\end{table}

Table~\ref{tab:accuracy} and~\ref{tab:map} show the evaluation results of our method and baselines on the KITTI-J and nuScenes-J datasets.
As shown, our PULSAR-Net can reconstruct $\geq$82\% of points on the nuScenes-J and $\geq$89\% of points on the KITTI-J when the simultaneous sensing is available (i.e., the number of simultaneous sensing lasers $\phi$ is $>1$), even though the naive average subtraction method can recover only $\leq$39\% of points on the nuScenes and $\leq$46\% of points on the KITTI-J. 
On the object detector-level mAP, our method can recover almost the same mAP ($\leq$65.52 mAP) as the benign one (69.87 mAP) without attack, while the naive average subtraction method can recover only up to 4.09\%, on the nuScenes-J dataset. On the KITTI-J dataset, while PULSAR-Net achieves $\geq$2 times higher performance than the naive average subtraction method, the performance is not as significant as the results in the nuScenes-J, and the naive average subtraction method also shows relatively high performance. We attribute the performance difference to the capabilities of the object detectors trained on each dataset, since the point-level accuracy is similar across both datasets. Since the KITTI dataset is dated compared to the nuScenes, we will mainly focus on the evaluation of the nuScenes-J in the rest of this paper.

For the lower performance of the naive average subtraction method, we find that it caused a large number of false positives because the attack pulses were not completely removed. The majority of attack pulses are dismissed by the naive average subtraction method, but it was not perfect, and the remaining attack pulses are still detected as points. Since the naive average subtraction method algorithm averages over multiple waveforms by design, legitimate laser pulses are inherently suppressed, making it difficult to distinguish true laser pulse peaks from false positives. These results indicate that using full-waveform data and simultaneous laser sensing information can provide significant benefits in denoising the impacts of jamming attacks. However, a straightforward average subtraction approach fails to handle the information adequately, and thus, the neural network approach of our PULSAR-Net plays a significant role in leveraging it. Neural DSP demonstrates some attack-mitigation capability by leveraging full-waveform information, but falls short of the performance achieved by PULSAR-Net. This is because Neural DSP is not designed for jamming attack scenarios, where the full waveform contains an excessive number of peaks that exceed the capacity of its current tokenization granularity. Finer-grained tokenization may alleviate this limitation, but remains an open question given the associated increases in computational and memory requirements.

\noindent\textbf{Transferability Analysis.} We evaluate the transferability of our PULSAR-Net trained on the different datasets to evaluate if our model learns LiDAR-agnostic physical principles rather than dataset-specific artifacts, enabling robust defense performance across diverse LiDAR systems. Table~\ref{tab:transfer_results} shows the model transferability analysis between KITTI-J and nuScenes-J datasets, on the point-level accuracy metric. The evaluation model is trained on one source dataset and evaluated in another target dataset. The number of simultaneous sensing lasers $\phi$ is 5, where the performance is already saturated. As shown, the PULSAR-Net trained on the nuScenes-J maintains high point-level accuracy even in the KITTI dataset. This indicates that the PULSAR-Net can acquire LiDAR-agnostic knowledge. However, it is still hard to deny the possibility that our PULSAR-Net may rely on some artifacts introduced by our attack simulation. We will further evaluate the transferability against the actual real-world attacks in~\S\ref{sec:real_world_eval}.

\vspace{-0.1in}
\subsection{Ablation and Efficiency Analysis} 
\vspace{-0.05in}
We conducted ablation and efficiency analysis of our PULSAR-Net. Based on the observations in~\S\ref{sec:main_result}, we evaluate our methods only on the nuScenes-J dataset with $\phi$=5. Further ablation study is in the supplementary material.

\begin{table}[t]
\centering
\small

\begin{minipage}[t]{0.49\textwidth}
\centering
\caption{
Model transferability analysis on the point-level accuracy metric.
The model is trained on the source dataset (row) and evaluated on the target dataset (column). The number of simultaneous sensing $\phi$ is 5.
}
\label{tab:transfer_results}
\vspace{-0.5em}
\setlength{\aboverulesep}{0pt}
\setlength{\belowrulesep}{0pt}
\begin{tabular*}{\linewidth}{@{\extracolsep{\fill}}llcc}
\toprule
& & \multicolumn{2}{c}{\textbf{Target}} \\
\cmidrule(lr){3-4}
& & nuScenes-J & KITTI-J \\
\midrule
\multirow{2}{*}{\textbf{Source}}
& nuScenes-J & \textbf{86.13} & 81.73 \\
& KITTI-J & 58.81 & \textbf{89.90} \\
\bottomrule
\end{tabular*}
\end{minipage}
\hfill
\begin{minipage}[t]{0.49\textwidth}
\centering
\caption{
Ablation study on axial spatial-temporal attention components.
}
\label{tab:attn_ablation}
\vspace{-0.5em}
\setlength{\aboverulesep}{0pt}
\setlength{\belowrulesep}{0pt}
\setlength{\tabcolsep}{0.8pt}
\begin{tabular*}{\linewidth}{
>{\centering\arraybackslash}p{0.09\linewidth}
>{\centering\arraybackslash}p{0.09\linewidth}
>{\centering\arraybackslash}p{0.09\linewidth}
|
>{\centering\arraybackslash}p{0.24\linewidth}
>{\centering\arraybackslash}p{0.24\linewidth}
>{\centering\arraybackslash}p{0.24\linewidth}
}
\toprule
\multicolumn{3}{c|}{Attention} & Point & Obj. & \\
W & H & D & Acc. & mAP & FPS \\
\midrule
 & & & 67.85 & 25.01 & 32.89 \\
$\checkmark$ & & & 82.98 & 62.05 & 31.94 \\
 & $\checkmark$ & & 80.34 & 62.50 & 32.25 \\
 & & $\checkmark$ & 81.93 & 62.55 & 32.36 \\
$\checkmark$ & $\checkmark$ & $\checkmark$ & \textbf{86.13} & \textbf{63.26} & 31.74 \\
\bottomrule
\end{tabular*}
\end{minipage}

\vspace{-0.1in}
\end{table}

\noindent\textbf{Impact of Axial Spatial-Temporal Attention.}
To evaluate the effectiveness of the axial spatial-temporal attention, we conduct an ablation study by applying attention along each individual axis and measuring its contribution to segmentation performance. As shown in Table \ref{tab:attn_ablation}, applying attention to a single axis provides marginal improvements, whereas the best performance is achieved when attention is jointly applied to all three axes. Since the FPS drop is negligible with axial attention enabled, our denoising model employs attention along all three axes of the waveform data. 

\noindent\textbf{Inference Efficiency.} PULSAR-Net achieves 31.74 FPS on an RTX 5090 even with the full size of the model used in our evaluation in~\S\ref{sec:main_result}. The perception module of autonomous driving typically works at 10-20 FPS~\cite{openpilot,apollo}, PULSAR-Net's throughput is on par with these industry standards. To further minimize computational burden, our defense can be paired with a lightweight attack detector such as~\cite{suzuki2025lab}, ensuring PULSAR-Net executes conditionally only under jamming attacks.

\vspace{-0.1in}
\section{Real-World Evaluation} \label{sec:real_world_eval}
\vspace{-0.05in}

\begin{figure}[t!]
\centering
\includegraphics[width=\linewidth]{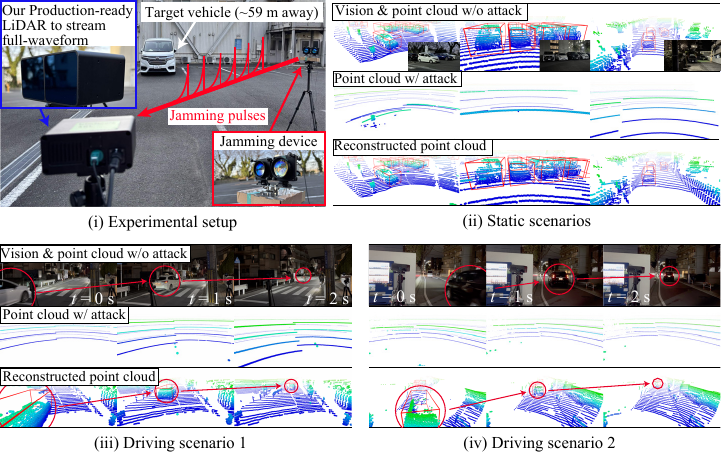}
\vspace{-0.3in}
\caption{
Overview of the real-world experiment and its results: (i) experimental setup of a jamming attack on our LiDAR developed with a major LiDAR supplier to stream full waveform, (ii) benign, attack-compromised, and reconstructed point clouds in a static scenario with a reference vision, and (iii-iv) 3 frames of camera frames and point clouds of benign, attack-compromised, and reconstructed point clouds  in driving scenarios
}
\label{fig:case_study}
\vspace{-0.15in}
\end{figure}

We conduct real-world evaluations to validate whether PULSAR-Net achieves sufficient generalizability when trained exclusively on simulated data. To this end, we develop a production-ready LiDAR sensor in collaboration with a major LiDAR manufacturer, capable of outputting full-waveform data while preserving the existing point cloud output, ensuring that our defense pipeline can operate in parallel with existing LiDAR perception systems in autonomous driving.

\noindent\textbf{Experimental Setups.}
The LiDAR (Fig. \ref{fig:case_study} (i)) we developed has a 90.24$^{\circ}$ horizontal and 26.3$^{\circ}$ vertical field of view (FoV), an angular resolution of 0.47$^{\circ}$ in both horizontal and vertical directions, and the number of simultaneous sensing lasers $\phi$ is 96. We launch jamming attacks against the LiDAR. We reproduce the HFR attack~\cite{sato2025on}, the current state-of-the-art LiDAR jamming attack.  We use the PULSAR-Net trained on the generated dataset based on nuScenes without any LiDAR-specific customization.
We conducted extensive experiments across eight scenarios, comprising three driving and five static sequences, totaling 700 frames (100 frames per sequence). To ensure a comprehensive evaluation, we manually annotated 3,106 vehicle instances. The ground-truth bounding boxes exhibit significant spatial diversity, with distances ranging from approximately 3.4 m to 59 m from the ego-vehicle. The LiDAR sensor was positioned at vehicle height, with a jamming device placed at various positions and angles relative to the sensor, as illustrated in Fig.~\ref{fig:case_study} (i). Our PULSAR-Net, trained on the nuScenes-J dataset, was evaluated using three representative 3D object detectors: PointPillars~\cite{lang2019pointpillars}, CenterPoint~\cite{yin2021center}, and PV-RCNN~\cite{shi2020pv}. These models were trained on the nuScenes dataset and implemented via the mmdetection3d framework~\cite{mmdetection3d}.

\begin{table}[t]
  \centering
  \caption{Quantitative results of the case study. We evaluate the mAP across three driving (moving) and four static scenarios. Note that benign (no attack) data is not available for moving scenarios.}
  \label{tab:case_study}

  \small
  \setlength{\aboverulesep}{0pt}
  \begin{tabular*}{\columnwidth}{@{\extracolsep{\fill}}ccccc}
    \toprule
    \textbf{Scenario} & \textbf{Condition} & \textbf{PointPillar} & \textbf{CenterPoint} & \textbf{PV-RCNN} \\
    \midrule

    \multirow{3}{*}{\shortstack[c]{\textbf{Static}\\(4 scenarios)}}
      & No Attack & 49.7 & 67.1 & 44.0 \\
      & No Defense & 0.10 & 0.00 & 0.00 \\
      & \textbf{Ours} & \textbf{49.8} & \textbf{69.4} & \textbf{49.8} \\
    \midrule

    \multirow{2}{*}{\shortstack[c]{\textbf{Moving}\\(3 scenarios)}}
      & No Defense & 0.70 & 0.20 & 0.00 \\
      & \textbf{Ours} & \textbf{50.7} & \textbf{63.1} & \textbf{53.8} \\
    \bottomrule
  \end{tabular*}
  \vspace{-0.18in}
\end{table}
\vspace{-0.1in}
\subsection{Results}

Table~\ref{tab:case_study} presents the evaluation results for the four static and three driving scenarios. Fig.~\ref{fig:case_study} illustrates the benign point cloud, the point cloud compromised by a jamming attack, and the point cloud reconstructed by PULSAR-Net.

\noindent\textbf{Static Scenarios}.
Table~\ref{tab:case_study} shows that jamming nearly collapses detection performance (mAP $\approx$0) in static scenes, while our reconstruction restores detector accuracy to near the benign level. Specifically, PULSAR-Net achieves 49.8/69.4/49.8 mAP for PointPillars/CenterPoint/PV-RCNN under attack, matching or slightly exceeding the no-attack baselines (49.7/67.1/44.0). With the best model (CenterPoint with IoU threshold=0.4), PULSAR-Net can reconstruct 91.7\% reconstruction accuracy for vehicles erased by attacks. This indicates that PULSAR-Net can not only reconstruct point clouds preserving the vehicle geometry needed for reliable detection, but also improve the detection performance with full-waveform information, as also observed in prior work~\cite{scheuble2025lidar}.

\noindent\textbf{Driving Scenarios}.
Table~\ref{tab:case_study} shows that jamming reduces detection to near zero in dynamic scenes (0.70/0.20/0.00 mAP), while PULSAR-Net restores strong performance (50.7/63.1/53.8) across all three detectors. This consistent recovery under motion indicates that the reconstruction preserves sufficient temporal and geometric cues for reliable detection even when vehicles are moving at up to 30.5 km/h. With the best model (CenterPoint with IoU threshold=0.4), PULSAR-Net achieves a reconstruction accuracy of 73.2\% for vehicles erased by attacks. We note that point clouds without attack are unavailable because we could not reliably replay driving scenarios in our testing facility, as shown in Fig.~\ref{fig:case_study} (iii-iv). We include demo videos in the supplementary material.

\vspace{-0.1in}
\section{Conclusion}
\vspace{-0.05in}
We establish PULSAR-Net, the first effective defense mechanism capable of reconstructing original point clouds from LiDAR sensing under jamming attacks. By leveraging previously unutilized intermediate full-waveform representations and simultaneous laser sensing in modern LiDAR systems, we demonstrated that it is possible to distinguish malicious attack signals from legitimate object reflections—a task previously considered theoretically impossible at the point cloud level. PULSAR-Net demonstrates over 82\% point-level recovery accuracy, representing more than 2 times improvement over a naive baseline approach. To validate its real-world viability, we collaborated with a major supplier to integrate PULSAR-Net into a production-ready LiDAR system. Despite being trained exclusively on synthetic data, our approach demonstrates high generalizability, achieving reconstruction success rates of 92\% in static scenarios and 73\% in driving scenarios. This work addresses a critical security vulnerability with direct implications for autonomous vehicle safety.

\appendix

\newpage
\section{Attck Demonstration Video}

We provide an attack demonstration video showing PULSAR-Net's performance against an actual LiDAR jamming attack, as described in Section 6. The video consists of driving scenarios (0–24s) and static scenarios (24–35s).
Object detection bounding boxes from CenterPoint are overlaid to show downstream task performance.

The video demonstrates three key findings:
\begin{itemize}
    \item \textbf{Attack severity}: The target vehicle completely disappears from the point cloud throughout the entire attack duration, confirming the jamming attack's persistent effectiveness across all frames
    \item \textbf{PULSAR-Net effectiveness}: Our method consistently recovers the vehicle structure in every frame during the reconstruction phase, demonstrating stable and reliable performance without intermittent failures  
    \item \textbf{Detection recovery}: Object detection fails in almost 100\% of frames under attack but achieves reconstruction success rate of 92\% in static scenarios and 73\% after PULSAR-Net reconstruction
\end{itemize}

This dynamic visualization effectively illustrates both the complete sensor blindness caused by jamming attacks and PULSAR-Net's robust recovery capability in real-world conditions.

\section{Limitations of Prior Point Cloud-Level Denoising Methods}

\begin{wrapfigure}{r}{0.5\textwidth} %
  \centering
  \vspace{-0.15in}
  \includegraphics[width=0.48\textwidth]{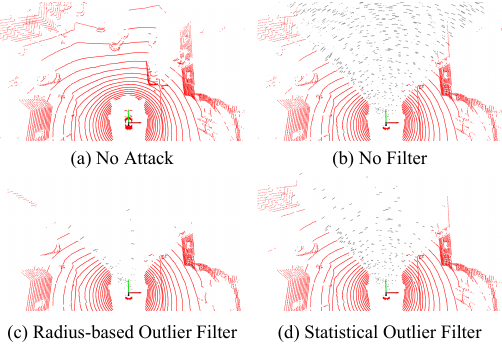} 
  \caption{Point cloud denoising methods fail under jamming attacks. Red points indicate valid structures (neighbors within 5cm), and gray points indicate isolated points.}
  \label{fig:point_cloud_denoisng}
  \vspace{-0.15in}
\end{wrapfigure}
As stated in Section 5.1, traditional point cloud denoising methods cannot recover objects under jamming attacks. Figure \ref{fig:point_cloud_denoisng} visualizes why these methods fundamentally fail.

We tested Radius Outlier Removal (ROR) and Statistical Outlier Removal (SOR)~\cite{Zhou2018} on jammed point clouds. The figure colors points red if neighbors exist within 5cm (valid structure) and gray otherwise (isolated noise). Panel (a) shows no attack with dense red structures representing real objects. Panel (b) reveals that jamming attacks completely destroy local geometric coherence—virtually no red points remain in the attacked region. Panels (c) and (d) show that both ROR and SOR filters remove some noise but cannot reconstruct any object structure.

This complete information loss at the point cloud level validates our approach of leveraging full-waveform representations, where attack and legitimate signals remain distinguishable by their temporal signatures and simultaneous sensing patterns, despite spatial randomization. Unlike point-based methods that operate on already-corrupted data, PULSAR-Net operates at the raw signal level, where legitimate reflections, though overwhelmed, still exist and can be separated from attack pulses.

\section{Implementation details of Neural DSP~\cite{scheuble2025lidar} baseline}

We reproduce Neural DSP~\cite{scheuble2025lidar} as an additional baseline for point cloud reconstruction. We reimplement its primary SWIN-based architecture with a Patch Tokenizer, while replacing the original Score and Offset classification heads with our segmentation head to ensure a fair comparison. As reported in the main paper, Neural DSP generally underperforms PULSAR-Net by several tens of percentage points in mAP and may collapse under certain settings, particularly at $\phi=225$ on the KITTI-J dataset.

The main bottleneck of Neural DSP arises from its low-resolution Patch Tokenization. Our target LiDAR waveforms have large spatial dimensions, e.g., $(32, 1800, 800)$, and directly tokenizing them into Transformer inputs of $(40, 128, 33)$ inevitably discards significant fine-grained waveform information. In particular, horizontal waveform correlations are heavily compressed, eliminating important cues for distinguishing real objects from malicious jamming signals. Although increasing the patch resolution could alleviate this problem, it would dramatically increase the token sequence length, leading to quadratic growth in memory usage and computational cost. For waveforms of $(32, 1800, 800)$, further increasing the tokenizer resolution becomes infeasible due to GPU memory constraints even on an RTX 5090. In contrast, PULSAR-Net explicitly focuses on efficient architectural design to capture fine-grained waveform structures while reducing computational overhead via DWSeparable 3D blocks, achieving a strong balance between performance and efficiency.

\section{Additional Ablation Study}
\noindent\textbf{Impact of Channel Dimension:}
The hidden channel dimension ($C$ in Fig.3 of the main paper) reflects the model’s representational capacity, with larger values generally improving its ability to capture complex waveform structures. We examine segmentation performance under different channel dimensions, and the results in Table~\ref{Table_Appendix_Channel_Dim} show steady improvement as $C$ increases from 16 to 32. Meanwhile, the FPS measured on an RTX 5090 GPU drops from 69.93 to 31.74,
it remains well above the threshold required for real-time processing.
Hence, we selected $C = 32$ for the denoising model as an optimal trade-off between accuracy and efficiency.

\noindent\textbf{Impact of Simultaneous Laser Sensing:}
We investigate the impact of simultaneous laser sensing. Our 3D convolution approach (\texttt{3D-CONV}) is designed to capture not only temporal information in the full waveform representation but also spatial information in simultaneous laser sensing. To highlight the impact of simultaneous laser sensing, we compare our approach with a 1D convolution along with the spatial axis. 

Table~\ref{Table_Appendix_CONV_Comp} shows the performance comparison bwtween \texttt{1D-CONV} and \texttt{3D-CONV}. As shown, \texttt{3D-CONV} shows significantly higher performance than \texttt{1D-CONV}, meaning modeling waveform shapes solely along the temporal axis is insufficient to differentiate object peaks from jamming attack's ones. Even though the full-waveform representation contains both legitimate and attack pulses, it is hard to remove the attack effects solely with the full-waveform representation. The simultaneous laser sensing plays an essential role in identifying the attack pulses as expected in Section 2.3. Furthermore, convolution along the spatial axis can achieve significant efficiency gains, as it acts as downsampling in the spatial dimensions. In Table~\ref{Table_Appendix_CONV_Comp}, it enables $>$70 times faster inference.

\begin{table}[htbp]
  \centering
  \small 
  \begin{minipage}[t]{0.49\textwidth}
    \centering
    \caption{Impact of channel dimensions ($C$) on PULSAR-Net performance.}
    \label{Table_Appendix_Channel_Dim}
    \begin{tabular}{c|cc|c}
    \toprule
    $C$ & \shortstack{Point Cloud\\level: Acc.} & \shortstack{Detection\\level: mAP} & FPS \\
    \midrule
    16 & 81.03 & 61.86 & \textbf{69.93} \\
    24 & 80.66 & 63.05 & 41.66 \\
    32 & \textbf{86.13} & \textbf{63.26} & 31.74 \\
    \bottomrule
    \end{tabular}
  \end{minipage}
  \hfill
  \begin{minipage}[t]{0.49\textwidth}
    \centering
    \caption{Comparison of performance between \texttt{1D-CONV} and \texttt{3D-CONV}.}
    \label{Table_Appendix_CONV_Comp}
    \begin{tabular}{c|cc|c}
    \toprule
    Model & \shortstack{Point Cloud\\level: Acc.} & \shortstack{Detection\\level: mAP} & FPS \\
    \midrule
    \texttt{1D-CONV} & 6.19 & 0.17 & 0.43 \\
    \texttt{3D-CONV} & \textbf{86.13} & \textbf{63.26} & \textbf{31.74} \\
    \bottomrule
    \end{tabular}
  \end{minipage}
\end{table}

\noindent\textbf{Waveform-Level Segmentation Metrics:}
While the main paper evaluates downstream performance, point-level recovery and object detection mAP, this supplementary section directly measures PULSAR-Net’s segmentation quality in the full-waveform domain. Since PULSAR-Net performs three-class segmentation (background, legitimate reflection, attack pulse) on the full-waveform tensor, evaluating segmentation accuracy at this level provides a more precise understanding of the model’s internal behavior.

We report IoU separately for legitimate object reflections (\emph{Object IoU}) and attack pulses (\emph{Attack IoU}). Object IoU measures how well PULSAR-Net preserves legitimate full-waveform returns, while Attack IoU measures how accurately it identifies malicious jamming pulses.

Table~\ref{Table_Appendix_CONV_Comp_Wide} shows that \texttt{3D-CONV} achieves substantially higher IoU for both legitimate object pulses and attack pulses than \texttt{1D-CONV}. This indicates that temporal information alone is insufficient and that incorporating spatial correlations from simultaneous laser sensing is essential for accurate waveform-level segmentation.

\begin{table}[htbp]
  \centering
  \begin{minipage}[t]{0.44\textwidth}
    \centering
    \caption{Waveform-level segmentation comparison between \texttt{1D-CONV} and \texttt{3D-CONV}. The \texttt{3D-CONV} achieves higher IoU for both legitimate object pulses and attack pulses.}
    \label{Table_Appendix_CONV_Comp_Wide}
    \small %
    \begin{tabularx}{\linewidth}{l|cc}
      \toprule
      \textbf{Model} & \textbf{Object IoU} & \textbf{Attack IoU} \\
      \midrule
      \texttt{1D-CONV} & 0.51 & 0.066 \\
      \texttt{3D-CONV} & \textbf{0.548} & \textbf{0.910} \\
      \bottomrule
    \end{tabularx}
  \end{minipage}
  \hfill %
  \begin{minipage}[t]{0.55\textwidth}
    \centering
    \caption{Dataset generation parameters}
    \label{tab:generation_params}
    \small
    \begin{tabular}{ll}
      \toprule
      \textbf{Parameter} & \textbf{Value} \\
      \midrule
      Attack interval $\delta_{\text{attack}}$ & 100 ns (10 MHz) \\
      Attack amplitude $\gamma_A$ & [3.0, 8.0] \\
      Timing jitter $\epsilon_k$ & $\mathcal{U}(-20, 20)$ ns \\
      \midrule
      Pulse scaling $\gamma$ (KITTI) & 12.0 \\
      Pulse scaling $\gamma$ (nuScenes) & 0.156 \\
      Time duration $T_{\text{max}}$ & 800 ns \\
      \bottomrule
    \end{tabular}
  \end{minipage}
\end{table}

\section{Full Experimental Results for Object Detection}

Table~\ref{tab:map} provides the comprehensive performance evaluation of PULSAR-Net and all baselines across all evaluated simultaneous sensing configurations on both nuScenes-J and KITTI-J datasets, revealing that PULSAR-Net consistently outperforms baseline methods across all configurations where simultaneous sensing is available ($\phi > 1$). Within this range, regardless of the number of simultaneous lasers, our method consistently maintains high $mAP$, demonstrating the robustness of our approach to leveraging simultaneous sensing information.

\begin{table}[h]
  \centering
  \caption{
    \textit{Object detector-level mAP} of PULSAR-Net (Ours) and baselines on the nuScenes-J and KITTI-J datasets. $\phi$ denotes the number of simultaneous sensing lasers.
  }
  \label{tab:map}

  \footnotesize
  \setlength{\aboverulesep}{0pt}
  \setlength{\belowrulesep}{0pt}
  \begin{tabular*}{\columnwidth}{@{\extracolsep{\fill}}clcc|cc}
    \toprule
    & & \multicolumn{2}{c|}{\textbf{nuScenes-J (mAP)}} 
      & \multicolumn{2}{c}{\textbf{KITTI-J (mAP)}} \\
    $\phi$ & \textbf{Method} 
      & PointPillars 
      & CenterPoint 
      & PointPillars 
      & PV-RCNN \\
    \midrule
    \multirow{4}{*}{1} 
     & No Defense & 0.00 & 0.00 & 0.00 & 0.00 \\
     & Naive Average Sub.  & - & - & - & - \\
     & Neural DSP~\cite{scheuble2025lidar} & 0.43 & 0.02 & 0.00 & 0.00 \\
     & \textbf{Ours} & \textbf{0.94} & \textbf{0.19} & \textbf{15.09} & \textbf{27.36} \\
    \midrule

    \multirow{4}{*}{4} 
     & No Defense & 0.00 & 0.00 & 0.00 & 0.00 \\
     & Naive Average Sub.  & 1.20 & 0.00 & 7.23 & 20.37 \\
     & Neural DSP~\cite{scheuble2025lidar} & 0.16 & 0.00 & 0.85 & 1.12 \\
     & \textbf{Ours} & \textbf{59.96} & \textbf{54.76} & \textbf{44.1} & \textbf{54.63} \\
    \midrule
    
    \multirow{4}{*}{5} 
     & No Defense & 0.00 & 0.00 & 0.00 & 0.00 \\
     & Naive Average Sub.  & 1.54 & 0.00 & 10.61 & 25.62 \\
     & Neural DSP~\cite{scheuble2025lidar} & 0.15 & 0.00 & 2.48 & 2.75 \\
     & \textbf{Ours} & \textbf{63.26} & \textbf{58.69} & \textbf{47.33} & \textbf{57.77} \\
    \midrule
    
    \multirow{4}{*}{10} 
     & No Defense & 0.00 & 0.00 & 0.00 & 0.00 \\
     & Naive Average Sub.  & 1.12 & 0.00 & 14.69 & 33.34 \\
     & Neural DSP~\cite{scheuble2025lidar} & 6.47 & 1.05 & 3.37 & 4.68 \\
     & \textbf{Ours} & \textbf{64.39} & \textbf{58.61} & \textbf{47.68} & \textbf{56.66} \\
    \midrule

    \multirow{4}{*}{25} 
     & No Defense & 0.00 & 0.00 & 0.00 & 0.00 \\
     & Naive Average Sub.  & 1.94 & 0.00 & 21.81 & 41.06 \\
     & Neural DSP~\cite{scheuble2025lidar} & 20.09 & 4.07 & 4.32 & 6.25 \\
     & \textbf{Ours} & \textbf{64.03} & \textbf{59.08} & \textbf{48.33} & \textbf{56.06} \\
    \midrule

    \multirow{4}{*}{55} 
     & No Defense & 0.00 & 0.00 & 0.00 & 0.00 \\
     & Naive Average Sub.  & 2.98 & 0.72 & 21.66 & 38.06 \\
     & Neural DSP~\cite{scheuble2025lidar} & 34.25 & 18.49 & 0.54 & 1.08 \\
     & \textbf{Ours} & \textbf{63.78} & \textbf{58.85} & \textbf{47.77} & \textbf{56.74} \\
    \midrule
     
    \multirow{4}{*}{110} 
     & No Defense & 0.00 & 0.00 & 0.00 & 0.00 \\
     & Naive Average Sub.  & 2.56 & 4.04 & 16.36 & 32.94 \\
     & Neural DSP~\cite{scheuble2025lidar} & 37.70 & 26.51 & 0.57 & 1.43 \\
     & \textbf{Ours} & \textbf{65.52} & \textbf{59.00} & \textbf{48.10} & \textbf{55.74} \\
    \midrule

    \multirow{4}{*}{225} 
     & No Defense & 0.00 & 0.00 & 0.00 & 0.00 \\
     & Naive Average Sub. & 4.09 & 9.11 & 0.14 & 2.34 \\
     & Neural DSP~\cite{scheuble2025lidar} & 31.05 & 22.49 & 0.34 & 0.63 \\
     & \textbf{Ours} & \textbf{64.15} & \textbf{59.25} & \textbf{48.95} & \textbf{56.67} \\
    \midrule

     & No Attack & 69.87 & 72.72 & 57.51 & 74.83 \\
    \bottomrule
  \end{tabular*}
  \vspace{-0.18in}
\end{table}

\section{Details of Dataset Generation}

We provide detailed the setup used in our full-waveform synthesis pipeline, designed to create realistic and challenging attack scenarios that encompass and extend beyond current state-of-the-art jamming attacks.

\subsection{Parameter Configuration}

Table~\ref{tab:generation_params} summarizes the key parameters used for dataset generation. These parameters were carefully selected to create a comprehensive threat model that surpasses existing jamming attacks.

To ensure that our reconstructed model remains robust under worst-case conditions, we design a jamming model whose strength surpasses that of the strongest known attacks (HFR~\cite{sato2024lidar} and A-HFR~\cite{sato2025on}). By incorporating substantially higher attack frequency, randomized amplitude modulation, and non-deterministic temporal jitter in our dataset generation pipeline, our threat model generalizes beyond existing HFR and A-HFR attacks and represents a strictly stronger adversarial setting. Specifically, the 10 MHz attack interval ($\delta_{\text{attack}} = 100$ ns) introduces far more pulses per LiDAR scan than existing attacks. %
Furthermore, the randomized amplitude $\gamma_A \in [3.0, 8.0]$ prevents simple threshold-based filtering while maintaining sufficient signal strength to overwhelm legitimate returns. Each pulse receives an independent random amplitude within this range, eliminating exploitable patterns. The temporal jitter $\epsilon_k \sim \mathcal{U}(-20, 20)$ ns applied to each pulse disrupts periodic patterns that could enable frequency-domain filtering. This 40 ns jitter window is sufficient to prevent coherent cancellation techniques while maintaining pulse integrity.

\subsection{LiDAR Signal Calibration}
\noindent\textbf{LiDAR Intensity Scaling:}
The pulse amplitude scaling factors $\gamma$ were calibrated separately for each dataset to account for differences in their LiDAR intensity representations: KITTI uses normalized intensity values in [0, 1], while nuScenes uses 8-bit values in [0, 255]. 
This scaled intensity directly represents the amplitude of the legitimate pulse within the full LiDAR waveform.

The selected scaling factors (12.0 for KITTI, 0.156 for nuScenes) satisfy two critical constraints: maintaining $\geq$95\% attack success rate while preserving sufficient legitimate pulse amplitude to prevent trivial intensity-based classification. Lower values would make defense unrealistically simple, while higher values would completely mask legitimate returns.

\noindent\textbf{Scan Duration of Full Waveform:}
Time duration of the scan $T_{\text{max}} = 800$ ns corresponds to the temporal dimension $D$, which translates to a maximum sensing range of 120 meters (using $d = c \cdot T / 2$). This range matches the valid detection distance of the HDL-64E used in KITTI, while exceeding the 100m range of the HDL-32E in nuScenes, ensuring complete coverage of all legitimate returns. 
Our U-Net-inspired architecture can generalize to longer temporal durations without retraining, as the convolutional operations naturally extend to different input sizes. This design choice ensures that PULSAR-Net can adapt to various LiDAR configurations with different maximum ranges, making it practical for deployment across diverse sensor specifications.

\section{Additional Discussions and Limitations}

\noindent\textbf{False Positive Analysis:}
One limitation of this work is the lack of a false-positive analysis in driving scenarios, as PULSAR-Net may introduce nonexistent ghost objects after denoising. While the impact of false positives is included in the mAP metric, this study does not validate PULSAR-Net's usability in actual driving scenarios. Meanwhile, we expect our PULSAR-Net to be used with an attack detection algorithm. As demonstrated in~\cite{suzuki2025lab}, state-of-the-art jamming attack detection achieves $\geq$90\% success rates even while the car is traveling at cruising speeds. 

\noindent\textbf{Practical Deployment Considerations:}
An attack-detection module should first detect a jamming attack, and PULSAR-Net should be activated only when such an attack is detected to avoid unnecessary computation during normal driving. However, detecting the presence of an attack does not itself ensure safe autonomous driving. Once an attack is detected, the vehicle still needs to maintain a safe trajectory; for example, slowing down or navigating to a secure area, especially during highway driving. In these situations, PULSAR-Net provides meaningful LiDAR sensing information even when the raw point cloud is unreliable, enabling safer decision-making under severe jamming attacks.

\noindent\textbf{Selective Attack in Simultaneous Sensing.}
PULSAR-Net assumes that attackers cannot inject different pulse patterns within the same simultaneous laser sensing. This assumption generally holds because it requires aiming multiple lasers with beam diameters $\leq$1 mm at a moving LiDAR from distances like 100 m. Detecting and targeting a LiDAR at such distances is already challenging; accurately directing multiple thin beams simultaneously is even less feasible in practice.

\noindent\textbf{Generality to Other LiDARs.}
PULSAR-Net requires full-waveform data and simultaneous laser sensing. Full-waveform data should be available in most LiDAR as it is directly received by their single-photon avalanche diodes. While simultaneous sensing may not be available, it is commonly used to achieve multiple vertical scan channels with no major technological barriers. 
For example, as shown in the Table~\ref{table:lidars}, modern multi-channel LiDAR systems already scan points simultaneously.
PULSAR-Net is thus applicable to most modern LiDARs, and we advocate that manufacturers should provide access to full-waveform data to enable more secure LiDAR applications. 
\begin{table}[t!]
    \small
    \centering
    \caption{Simultaneous laser sensing capabilities of commercial LiDAR systems. $\phi$ is the number of simultaneous sensing lasers.}
    \setlength{\tabcolsep}{2pt}
    \setlength{\aboverulesep}{0pt}
    \setlength{\belowrulesep}{0pt}
    \label{table:lidars}
        \begin{tabular}{cccc}
            \toprule
            & \footnotesize Robosense M1P~\cite{M1P} & \footnotesize Hesai AT 128P~\cite{AT128P} & \footnotesize Ouster VLS128~\cite{VLS128} \\
            &
            \includegraphics[width=0.09\columnwidth]{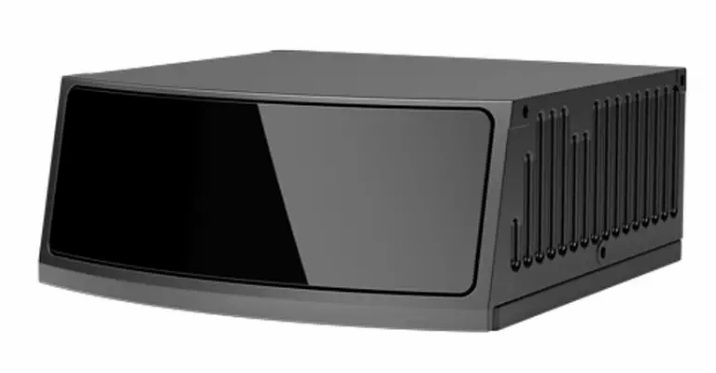} &
            \includegraphics[width=0.13\columnwidth]{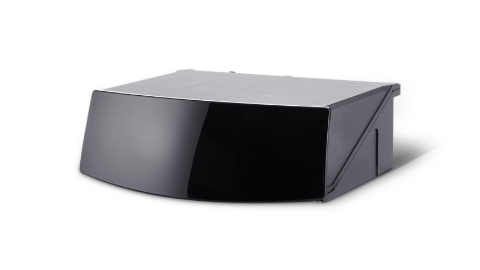} &
            \includegraphics[width=0.07\columnwidth]{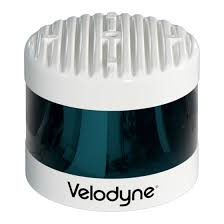} \\
            \midrule
            $\phi$ &
            32 & 16 & 8 \\
            \bottomrule
        \end{tabular}
        \vspace{-0.1in}
\end{table}

\newpage
{
    \small
    \bibliographystyle{splncs04}
    \bibliography{main}
}

{
    \small
    \bibliographystyle{splncs04}
    \bibliography{main}
}

\end{document}